\def\year{2022}\relax
\documentclass[letterpaper]{article} 
\usepackage{aaai22}  
\nocopyright
\usepackage{times}  
\usepackage{helvet}  
\usepackage{courier}  
\usepackage[hyphens]{url}  
\usepackage{graphicx} 
\usepackage{natbib}  
\usepackage[justification=centering]{caption} 
\DeclareCaptionStyle{ruled}{labelfont=normalfont,labelsep=colon,strut=off} 
\usepackage{algorithm}
\usepackage{algorithmic}
\usepackage{newfloat}
\usepackage{listings}
\floatstyle{ruled}
\newfloat{listing}{tb}{lst}{}
\floatname{listing}{Listing}

\usepackage{amsmath}
\usepackage{amsfonts}
\title{Protecting Anonymous Speech: A Generative Adversarial Network Methodology for Removing Stylistic Indicators in Text}
\author {
    Rishi Balakrishnan,
    Stephen Sloan,
    Anil Aswani
}
\affiliations {
    University of California, Berkeley \\
    rishibalakrishnan@berkeley.edu, stephen\_sloan@berkeley.edu, aaswani@berkeley.edu
}
\begin{document}
\maketitle
\begin{quote}
\begin{abstract}
With Internet users constantly leaving a trail of text, whether through blogs, emails, or social media posts, the ability to write and protest anonymously is being eroded because artificial intelligence, when given a sample of previous work, can match text with its author out of hundreds of possible candidates. Existing approaches to authorship anonymization, also known as authorship obfuscation, often focus on protecting binary demographic attributes rather than identity as a whole. Even those that do focus on obfuscating identity require manual feedback, lose the coherence of the original sentence, or only perform well given a limited subset of authors. In this paper, we develop a new approach to authorship anonymization by constructing a generative adversarial network that protects identity and optimizes for three different losses corresponding to anonymity, fluency, and content preservation. Our fully automatic method achieves comparable results to other methods in terms of content preservation and fluency, but greatly outperforms baselines in regards to anonymization. Moreover, our approach is able to generalize well to an open-set context and anonymize sentences from authors it has not encountered before.
\end{abstract}
\end{quote}

\section{Introduction}
Stylometric authorship attribution attempts to identify the author or attributes of the author of a given piece of text. Stylometry uses features such as frequency of words or words per sentence, and first found success in authorship attribution for the Federalist Papers \cite{c:28}. Stylometric methods originally had to be done manually, but can now be automated using machine learning. Stylometric analysis was also successfully used to help identify J.K. Rowling as the author of \emph{A Cuckoo’s Calling}, which she wrote under a pseudonym \cite{c:23}. Similar methods were used to identify some of Shakespeare’s coauthors \cite{c:24}. At the larger scale, \citet{c:29} were able to use stylometry to successfully identify the author of a piece of text from a set of 100,000 authors with 20\% accuracy. \citet{c:30} were even able to identify the author of code from their coding styles. Furthermore, \citet{c:49} show that it is possible to identify when an author is manually trying to change their own writing style, meaning this is not a possible method for evading authorship attribution in the long run.

The success of stylometric authorship attribution methods poses a problem for authors who wish to publish anonymously, damaging free speech and civil liberties \cite{c:26}. The increased usage of services like Tor in oppressive regimes because of the fear of identification demonstrates the need for privacy in political discourse \cite{c:27}. Even in democracies, political dissidents may depend on anonymity to expose government malfeasance to the public. A recent example of this is a 2018 memo published by an (anonymous at the time of publication) official in the Trump administration \cite{c:25}. After publication, leading experts in linguistics received several requests to use stylometric techniques to identify the author \cite{c:61}. An FBI report details ``a growing need to identify writers not by their written script, but by analysis of the typed content'' \cite{c:36}, and work like \cite{c:37} argues that stylometry should be used as a biometric trait, especially in combination with other data like keystrokes. Taken together, the usage of stylometric techniques to identify authorship will only increase, which creates a need for systems to protect whistleblowers and political dissidents through anonymizing their written content.

\textbf{Our Contributions} We make two major contributions to the field of authorship obfuscation: 1) We develop an adversarial method where the discriminator better captures style, enabling better anonymization than previous methods and 2) To the best of our knowledge, we provide the first authorship-based anonymization framework that aims to obfuscate authorship in an open-set context. Our experiments prove that  empirically our model generalizes well to provide anonymity for unseen authors against unseen classifiers. 

\section{Related Work}
\textbf{Generative Adversarial Networks}
A Generative Adversarial Network (GAN) is a machine learning framework that usually has two models competing against each other \cite{c:1}. In the traditional formulation, the generator $G$ takes in random (usually Gaussian) noise $z \sim Normal(0, 1)$ and outputs $G(z)$. The discriminator $D$ learns to distinguish between real samples $x$ and generated samples $G(z)$ and the generator learns to output samples as close to the real data distribution as possible, with the generator and discriminator playing a min-max game of the form
\begin{equation}
    \min \max \mathbb{E}_{z}[\log D(G(z))] + \mathbb{E}_{x}\log (1-D(x))]
\end{equation}
However, adversarial training has proven tricky with common issues being mode collapse and non-convergence \cite{c:11}. Mode collapse occurs when the generator finds a particular output for which the discriminator misclassifies samples and exclusively produces that output. The most prominent methods proposed to solve these training issues include the Wasserstein GAN, which changes the traditional GAN objective function from Jensen-Shannon divergence to Earth Mover’s Distance \cite{c:12}
\begin{equation}
    \min \max \mathbb{E}_{z}[D(G(z))] - \mathbb{E}_{x}[D(x)]
\end{equation}
These GAN’s use either gradient penalties \cite{c:44} or weight-clipping \cite{c:12} to impose a 1-Lipschitz constraint on the discriminator training, which helps the generator smoothly learn the distribution of the real data. Due to this and other training advancements, GAN’s have found widespread adoption within computer vision, including applications in conditional image generation \cite{c:2}, 3D object modeling \cite{c:3}, and image style transfer \cite{c:4}. 

Unfortunately, the discrete nature of text complicates GAN usage within natural language processing tasks. Selecting discrete text tokens from the generator output makes backpropagating gradients from the discriminator to the generator difficult. Because of this, gradient-estimation methods like REINFORCE \cite{c:5} and Gumbel Softmax \cite{c:6,c:7} have gained popularity. REINFORCE combines techniques inspired from reinforcement learning with GAN's, using the policy-gradient technique to train the generator. The REINFORCE method has been used to train text GAN’s including LeakGAN \cite{c:8}, MaskGAN \cite{c:9}, and SeqGAN \cite{c:10}. In contrast, the Gumbel Softmax technique adds noise sampled from the Gumbel distribution to the output logits and then applies a softmax operation with temperature $\tau$ to the result. As $\tau$ decreases, the softmax output becomes more one-hot. This method offers a continuous relaxation of the discrete probabilities output by a model, which allows gradients to backpropagate through a discrete sample.

\textbf{Siamese Networks}
Siamese discriminators first found success in signature verification \cite{c:45}, where they took in images of two signatures and determined whether they originated from the same person. They have later been applied to various tasks, including facial recognition \cite{c:46}, object tracking \cite{c:47}, and one shot image recognition \cite{c:48}. Siamese networks are usually structured as two subnets which share the same parameters and weight updates. When given two inputs, each subnet outputs an embedding for its respective input. A distance metric (e.g., L2 distance or cosine similarity) then compares the two embeddings to determine whether the two inputs are from the same class or not. As a result, the network learns to cluster similar samples close together and dissimilar inputs far apart. Siamese networks offer powerful opportunities for generalizable classification, because when the network is used in a classification setting, given an example of a new target class, the network can determine whether inputs belong to that class or not.

\textbf{Authorship Attribution and Verification}
The field of authorship attribution stretches back to before the advent of most machine learning methods. Early work aimed to use statistical features of written text, including word and n-gram frequency, sentence length, punctuation, etc. to identify the author \cite{c:17}. However, modern natural language processing techniques have shifted authorship attribution methods to take in raw text rather than pre-computed features of the raw text. Recent work uses character-level and word-level convolutional neural networks (CNN’s) \cite{c:14} or multi-headed recurrent neural networks (RNN’s) \cite{c:20} for authorship attribution, which achieve significant improvements over non-deep learning methods including support vector machines (SVM’s) and linear discriminant anaysis (LDA) modelling. 

An alternate approach to authorship attribution is authorship verification. This approach instead aims to embed the stylistic components of text into a vector and determine whether two pieces of text are written by the same author. Therefore, it can also be used in a classification setting by assigning authorship based on similarity between the sampled text and text with known authorship. For example, \cite{c:19} use Siamese networks to verify if two pieces of text share the same author. 

\begin{figure*}
    \centering
    \includegraphics[width=15cm]{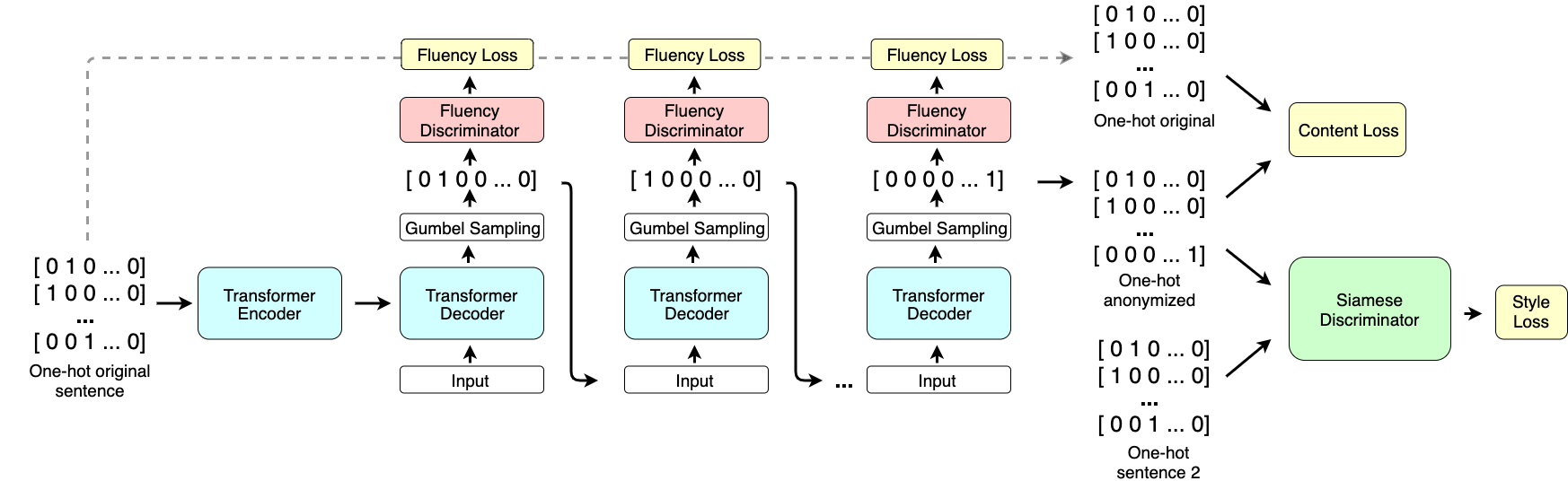}
    \caption{Network Diagram}
    \label{Figure 1}
\end{figure*}

\textbf{Authorship Anonymization}
Initial efforts to conceal authorship largely focused on rule-based and manual interventions to modify authors’ styles. Anonymouth \cite{c:17} was an open-source platform released that suggests changes to users based on features like word and character counts, average syllables per word, sentence length, and readability scores. However, these suggested transformations are heavily dependent on the dataset the tool uses, which limits the tool’s usefulness. A simple approach is round-trip translation \cite{c:19,c:62}, where, for example, an English input text is translated into several other languages then finally translated back into English. However, these methods depend on inaccuracies or distortions introduced by the translation software, and even with distortions, authorship attribution systems were able to identify the original authors with high accuracy \cite{c:18}. 

Modern work approaches anonymization in one of two ways -- imitation or obfuscation. Imitation conceals an author’s identity by mimicking another author’s style, while obfuscation masks stylistic indicators as a whole (i.e., to reduce the accuracy of a classifier to chance level). Anonymization work related to imitation mirrors work in the related field of text style transfer. The authors of \cite{c:20} design a system to hide binary demographic attributes, like age or gender, through imitation. Their model consists of an encoder-decoder network to generate text, back-translation loss to preserve content, and style classifiers to distinguish between real and generated sentences. This system, however, protects only one binary demographic attribute rather than the whole identity, and can only protect that binary attribute if it had been trained to do so. As a result, obfuscation approaches to anonymization should be preferred. \citet{c:21} use genetic algorithms to anonymize sentences, iteratively changing sentences until a classifier is fooled. This approach, however, only works against a known adversary. Deep learning approaches to authorship obfuscation include work such as \cite{c:22} which designs an authorship anonymization system using techniques from differential privacy. \citet{c:64} use a gradient reversal layer between sentence embedding and authorship classifier in order to make text style-invariant. They demonstrate the promise of machine learning methods in anonymization by obfuscation, but similar to \cite{c:21}, the method is only trained on a limited set of 5 authors.

\section{Methodology}
\textbf{Our Approach}
The stylometric approach to authorship attribution depends on the assumption that style can be modeled as a probability distribution. Indeed, some work in linguistics defines style as intentional overuse of certain linguistic features and norms \cite{c:34} in expressing the same content. This means we can define style as the way that individual authors choose to express the same content, and that there exists a style-neutral expression of that same content -- the one that most closely adheres to linguistic norms. Thus, the purpose of any obfuscation system is to learn a many-to-one mapping that transforms a sentence from an author's style to the style-neutral version while retaining the same content.

A survey of previous work within stylometry and authorship attribution explored the use of different features in their ability to predict authorship, including word and n-gram frequency, sentence length and structure, syntax, parts-of-speech usage, and rhythm, among others \cite{c:35}. The survey concluded that n-gram, character, and word frequency were the features most salient to authorship attribution, because these features (n-grams especially) automatically capture many other relevant parts of text, including punctuation and morphology. These observations shape our approach in two ways: 1) we focus primarily on word and n-gram level changes (instead of features like sentence structure) to thwart authorship attribution systems and 2) we anonymize text on a sentence-level rather than on the level of a paragraph or document because the relationship between different sentences and paragraphs is viewed as secondary.

\textbf{Network Structure}
The network we employ in this model is structured according to Figure 1. There are two components to the Transformer language model. The Transformer encoder first captures the content of a sentence and embeds it into a vector, and then the decoder auto-regressively generates output tokens in a style-neutral fashion. As well as being used in the final anonymized input, each generated token is fed into a fluency discriminator which creates the fluency loss by judging how likely each token is to be from a real sentence. Text is generated until a maximum sequence length, and is then fed into the Siamese discriminator along with a second sentence, and the discriminator determines how likely it is that both sentences are from the same author. This likelihood becomes the style loss for the generator. Finally, both the anonymized and original sentences are fed into the mechanism that we use to generate the content loss, which we discuss below.

\textbf{Style Discriminator}
We employ a Siamese network as the primary discriminator for the GAN. The function of the discriminator is authorship verification -- taking in two pieces of text and identifying whether they are by the same author. We select an authorship-verification approach for the discriminator over an authorship-classification approach for two reasons. First, anonymization with a classification model as the adversary may not generalize to authors outside the training set. \citet{c:38} uses the analogy of fruit classification to explain: color helps differentiate apples from bananas, but if a model based on color is then used to distinguish pears and lemons, it will fail. Instead, if the generator learns to anonymize sentences from a discriminator that embeds the sentence style as a whole, it will likely be able to generalize better. Second, because this approach requires capturing style as a whole rather than learning features specific to authors, the discriminator’s task becomes more difficult. This stabilizes GAN training -- an overly easy task for the discriminator leads to high discriminator accuracy and vanishing gradients, preventing the generator from learning \cite{c:39}. 

For the model architecture itself, we use stacked ResNet blocks with convolutional layers \cite{c:43}. CNN-based architectures have proven effective at natural language tasks including sentiment analysis and question classification \cite{c:42}. The authors of \cite{c:15} design a Siamese network for authorship classification and verification using stacked convolutional layers, outperforming previous authorship verification methods and achieving comparable results on n-way classification to other baselines. The convolutional layers of the discriminator as well as the residual connections within each ResNet block allow easy backpropagation of gradients and prevent vanishing gradients.

We structure the GAN network as a Wasserstein GAN with Gradient Penalty (WGAN-GP). The central innovation of the WGAN is conceptualizing the discriminator as a 1-Lipschitz function, which has gradients whose norms are less than 1, which helps stabilize training by preventing the discriminator from learning too quickly. The original paper with Wasserstein GAN uses weight clipping to enforce the Lipschitz constraint, but work presented in \cite{c:44} that improves WGAN performance uses a gradient penalty to softly enforce that gradients are near 1. To calculate the gradient penalty, the original input is interpolated with the anonymized output 
\[
    \hat{x} = a_{int}x + (1-a_{int})G(x)
\]
for some $a_{int} \in [0, 1]$ and then fed into the discriminator to calculate the norm of the discriminator gradients. As a result, the loss function for the style discriminator is 
\begin{equation}
\begin{split}
    L_{sty} &= - \mathbb{E}_{x_1, x_2 \sim p(a_i)}[D_{sty}(G(x_1), x_2)] \\ 
    &+ \mathbb{E}_{x_1 \sim a_i, x_2 \sim a_j, j \neq i}[D_{sty}(x_1, x_2)] \\
    &+ \lambda(||\nabla_{\hat{x}} D_{sty}(\hat{x})||_{2} - 1)^2
\end{split}
\end{equation}
\textbf{Fluency}
The simplest method for the generator to fool the discriminator is to output broken sentences whose style cannot be deciphered by the discriminator. To ensure the fluency of generator outputs, we employ a discriminator to distinguish between real and generated sentences. However, as \cite{c:40} explain, the error signal from using a classifier to evaluate the fluency of the entire sentence is often too weak to force the generator to create fluent outputs. Instead, their approach uses a language model to provide token-level feedback at every step of the generator output. While the task in \cite{c:40} is text style transfer, we draw inspiration from their method and feed the output of the generator at each time step into a one-layer Gated Recurrent Unit (GRU) RNN discriminator, generating a score for every word. Given generator pre-training (which is explained shortly), the initial outputs of the generator during training are fluent. As a result, the purpose of the fluency loss is not to provide a distribution that the generator has to learn but rather to guide the generator to stay fluent throughout training. As a result, we do not impose the gradient penalty on the fluency discriminator during training. Therefore, the loss function for the training of the fluency discriminator is 
\[L_{fl} = - \mathbb{E}_{x}[D_{fl}(G(x))] + \mathbb{E}_{x}[D_{fl}(x)] \]

\textbf{Content}
\begin{figure}
    \centering
    \includegraphics[width=8cm]{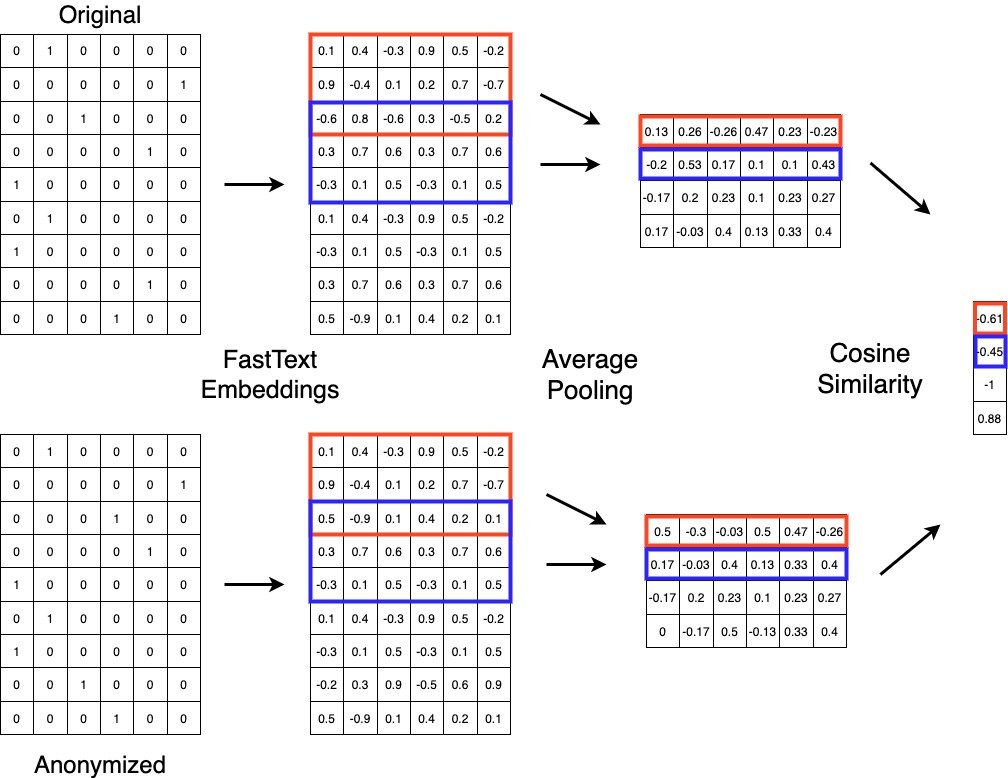}
    \caption{Content Loss Generation}
    \label{Figure 2}
\end{figure}
Similar to the smoothing loss, our content preservation approach provides near-token level feedback to the generator, as detailed in Figure 2. We embed both the input and the output of the generator using FastText embeddings, and then use an Average Pooling layer to generate vectors for both embeddings given a pool size and stride. These vectors are then compared using cosine similarity and the mean is taken to generate a score of how consistent the content of the generated sentence is with the original. We denote the final content loss as $L_{con}$, and because higher values for cosine similarity are better, we minimize $-L_{con}$. In practice, we note that providing this token or near-token level feedback to the generator for both fluency and content preservation also helps prevent mode collapse.

\textbf{Generator}
For the generator, we use the vanilla Transformer model with self-attention as discussed in \cite{c:31}. Transformers have demonstrated the ability to capture long-term dependencies better than their alternatives like RNN’s. The Transformer model consists of both an encoder and decoder, which embeds the original sentence and generates tokens respectively. Both the encoder and decoder have 4 stacked Transformer blocks, which consist of multi-head attention -- which determines the most relevant parts of the original sentence -- as well as fully-connected layers and layer normalization. Additionally, the decoder employs self-attention, looking at previously generated tokens to determine the next one. Each Transformer block also employs residual connections to help gradients flow easily.

In text GAN’s, generators are often pre-trained using maximum likelihood estimation (MLE) and teacher forcing. Teacher forcing feeds the model with the ground truth sequence at every time step instead of using previous predictions. While effective during training, this creates a discrepancy between training and inference time when the Transformer must generate the entire sentence itself \cite{c:32}, and this discrepancy degrades performance at inference time. To correct this, we employ scheduled sampling as described in \cite{c:33}, where instead of only feeding in the ground truth sequence, certain time steps sample the previous output of the generator instead, with the sampling rate increasing during training. 

For the adversarial training, we combine weighted style, fluency, and content losses to create the objective function of the generator
\begin{equation}
\begin{split}
    L_{G} &= a_{sty} \mathbb{E}_{x_1, x_2 \sim a_i}[D_{sty}(G(x_1), x_2)] \\ &+ a_{fl} \mathbb{E}_{x}[D_{fl}(G(x))] - a_{con}L_{con}
\end{split}
\end{equation}

\textbf{Training}
To ensure that the tokens from the generator are still differentiable, we employ the Gumbel Softmax technique instead of REINFORCE, because the REINFORCE method generates gradients that have much higher variance than those created by the Gumbel Softmax \cite{c:6}. Specifically, we use the Gumbel-Softmax straight-through trick, where the token with the maximum probability is selected in the forward pass, but the gradients from the Gumbel Softmax are used in the backward pass.

Additionally, because learning word similarities and associations becomes difficult given a limited corpus and our approach to anonymization largely depends on rephrasing and word replacement, FastText embeddings pretrained on the Common Crawl dataset \cite{c:41} are used for the model and kept static throughout training. Because of the importance of n-grams in author identification, we also employ Wordpiece Tokenization \cite{c:51}, which breaks down a word into the most common subword units and ensures that every word can be tokenized. This also means that the FastText embeddings used are sub-word rather than word embeddings.

For the model training, we follow the Two Time Update Rule in \cite{c:50}, which proves that when using the Adam optimizer with a higher learning rate for the discriminator than the generator, the GAN is likely to converge to a local Nash equilibrium. \cite{c:50} recommends that for a WGAN-GP model trained on text, the discriminator and generator should be trained in an alternating fashion, with one epoch for the discriminator and then one epoch for the generator. We also note that in practice, performing model updates for the fluency discriminator as well as the generator in the same step helps catch mode collapse early, so weight updates to the style discriminator occur simultaneously with weight updates to the generator. 


\begin{table*}
\begin{center}
\begin{tabular}{c | c  c | c | c  c  c} 
& \multicolumn{2}{c|}{Content Preservation} & Fluency & \multicolumn{3}{c}{Anonymity}\\ 
  & BLEU$\uparrow$ & BERTScore$\uparrow$ & PPL$\downarrow$ & RNN$\downarrow$ & CNN$\downarrow$ & Transformer$\downarrow$\\ 
  \hline \hline
 Original & N/A & N/A & 5.71$\pm$0.009 & 0.42$\pm$0.018 & 0.47$\pm$0.014 & 0.49$\pm$0.015\\
 WordNet & 0.87$\pm$0.001 & \textbf{0.99$\pm$0.000} & 6.73$\pm$0.012 & 0.35$\pm$0.015 & 0.36$\pm$0.011 & 0.38$\pm$0.013\\
  Roundtrip &0.20$\pm$0.001 & 0.93$\pm$0.000 & 6.27$\pm$0.017 & 0.30$\pm$0.016 & 0.27$\pm$0.012 & 0.32$\pm$0.014\\
 Mutant-X & \textbf{0.93$\pm$0.003} & \textbf{0.99$\pm$0.000} & \textbf{6.17$\pm$0.038} & 0.36$\pm$0.018 & 0.38$\pm$0.017 & 0.39$\pm$0.016 \\
 Ours & 0.61$\pm$0.002 & 0.93$\pm$0.000 & 6.98$\pm$0.011 & \textbf{0.18$\pm$0.014} & \textbf{0.19$\pm$0.012} & \textbf{0.21$\pm$0.015} \\
\end{tabular}
\end{center}
\caption{Results for closed-set evaluation are reported as mean $\pm$ standard error. $\uparrow$ and $\downarrow$ represent that a higher and lower score are better, respectively. The RNN, CNN, and Transformer columns report the F1 score.}
\label{table:1}
\end{table*}

\section{Experiments}
\textbf{Dataset}
In line with the goal of our system being used primarily to protect the free speech of journalists and whistleblowers, the dataset we use within this paper is the Reuters C50 dataset \cite{c:53}, a subset of the much larger Reuters Corpus V1 dataset \cite{c:52}. To create the Reuters C50 dataset, articles not related to corporate or industrial news are filtered out to reduce the impact of topic on author identification. Then, 50 articles from the top 50 news authors (with respect to number of articles written) are selected. We split each article into sentences and remove sentences with fewer than 10 characters to ensure meaningful sentences. Punctuation is kept when pre-processing because it can indicate style, but to reduce the impact of proper nouns on identifying authorship, we use the Spacy natural language processing library \cite{c:54} to identify geopolitical entities, people, organizations, and locations and replace them with “[GPE]”, “[PERSON]”, “[ORG]”, and “[LOC]” tokens respectively. To prepare the dataset for the Siamese discriminator, we create sentence pairs with 2 same-author sentence pairs and 2 different-author sentence pairs for each sentence in the original dataset.

We use the dataset in two contexts. First is a closed-set context where all authors are present in the training, validation, and test sets to evaluate how well our system anonymizes sentences from authors it has encountered. In this first context We use a training-validation-testing set split of 75\%-15\%-15\% of the articles of each author. The second is an open-set context where we test how well our system anonymizes sentences from authors it hasn't encountered. We use 30 authors for the training set, 10 authors for the validation set, and 10 authors for the testing set.

\textbf{Implementation Details}
We implement our framework using TensorFlow, training on two NVIDIA Tesla P-100 GPU's for around 8 hours for both the open-set and closed-set models. We performed a hyper-parameter sweep for the values of $a_{sty}$, $a_{fl}$, and $a_{con}$ using combinations of integer values in the interval $[1, 10]$ before settling on $a_{sty}=1$, $a_{fl}=7$, $a_{con}=6$ based on perplexity, content similarity, and anonymization results on a validation dataset. We set $\tau=1$ and $a_{int}=.5$ initially and find that they work well in practice. Following the recommendations of \cite{c:50}, we use Adam optimizers for all models with $\beta_1=0.5$ and $\beta_2=0.9$ and initially set the learning rate of the discriminators to .0003 and the generator to $.0001$. We then experimented with values close to those initial values, eventually using a learning rate of .00008 for the generator and .0004 for the style and fluency discriminators. We find the discriminator learns most effectively for small $\lambda$ values, and best when $\lambda=.05$. We train the generator and discriminators for 15,000 weight updates in the closed-set context and 14,000 weight updates in the open-set context.

\textbf{Evaluation Metrics}
There are three points of evaluation for the output of the model, each corresponding to one of the three different losses for the generator. 

\textit{Anonymity}: To evaluate the effectiveness of the model in anonymizing text inputs, we evaluate how well different anonymization methods decrease the F1 score of classifiers trained on the training set. Classifiers with different architectures may learn to classify sentences using different features, and we want to ensure that the anonymization is robust to various adversaries. We employ three different classifiers: 
\begin{itemize}
    \item A GRU RNN followed by two fully connected layers
    \item A convolutional model with three convolutional layers with stride 1 and pool size 3, 4, and 5 with max pooling followed by two fully connected layers
    \item A multi-head attention based Transformer encoder model then connected to two fully connected layers
\end{itemize}
For open-set evaluation, while our model has not seen authors in the test set, we evaluate the model against adversaries that have. For better evaluation, we split the sentences in the test set into 70\%-30\% training-validation subsets, and train the classifiers on the training subset. The efficacy of anonymization is then evaluated on the validation subset. This ensures that changes in classification of the anonymized sentences happen because the sentences have been anonymized rather than the classifiers overfitting to their training data and being overly sensitive to \textbf{any} change. We provide the average F1 scores and standard error for classifications to judge how well our system anonymizes each author. Accuracy closely resembled the F1 score in our results, and so we do not report it. 
\begin{table*}
\begin{center}
\begin{tabular}{c | c  c | c | c  c  c} 
\ & \multicolumn{2}{c|}{Content Preservation} & Fluency & \multicolumn{3}{c}{Anonymity}\\ 
  & BLEU$\uparrow$ & BERTScore$\uparrow$ & PPL$\downarrow$ & RNN$\downarrow$ & CNN$\downarrow$ & Transformer$\downarrow$\\ 
 \hline \hline
 Original & N/A & N/A & 5.92$\pm$0.020 & 0.93$\pm$0.005 & 0.94$\pm$ 0.004 & 0.92$\pm$0.006 \\
 Round-trip & 0.23$\pm$0.002 & \textbf{0.93$\pm$0.001} & \textbf{6.10$\pm$0.034} & 0.64$\pm$0.025 & \textbf{0.73$\pm$0.020} & 0.69$\pm$0.023 \\
 Ours & \textbf{0.64$\pm$0.004} & \textbf{0.93$\pm$0.001} & 6.95$\pm$0.023 & \textbf{0.63$\pm$0.024} & 0.76$\pm$0.024 & \textbf{0.68$\pm$0.025} \\
 \end{tabular}
\end{center}
\caption{Results for open-set evaluation are reported as mean $\pm$ standard error. $\uparrow$ and $\downarrow$ represent that a higher and lower score are better, respectively. The RNN, CNN, and Transformer columns report the F1 score. Results for WordNet or Mutant-X are not provided because these methods cannot be used in an open-set context.}
\label{table:2}
\end{table*}

\textit{Fluency}: To measure the fluency of the outputs, we use the GPT-2 model provided by the HuggingFace library \cite{c:55}. We evaluate the perplexity score (PPL) of the model on the anonymized sentences, with the intuition being that for coherent, fluent sentences, the next word is much easier to predict which leads to a lower PPL overall. We provide the mean and standard error of the PPL scores. 

\textit{Content Preservation}: To measure how well anonymization preserves the content of the original sentence, we use two different metrics. The first is the BLEU score \cite{c:56}, a standard method of evaluation for machine translation and related tasks that measures n-gram similarity between the generated sentence and the reference sentence (i.e., the sentence before obfuscation). However, the BLEU score has been criticised: \citet{c:57} find that the BLEU score often correlates poorly with human judgement. \citet{c:58} explains that the BLEU score should not be used outside of machine translation tasks, and \cite{c:59} argues it is inadequate to evaluate text simplification. The reason for these issues is because BLEU evaluates content preservation by n-gram similarity, not by the meaning of the sentences. As a result, BLEU is imperfect to evaluate authorship obfuscation, because as we mention in the methodology section, authorship obfuscation methods should replace low frequency n-grams with higher frequency n-grams. We include the BLEU score because it is standard, but supplement it with BERTScore evaluations \cite{c:60}. The BERTScore uses large language models to evaluate the content preservation of a sentence given a reference, and has shown to correlate well with human judgement. We provide the mean and standard error of both the BLEU scores and BERTScores.

\textbf{Baseline Methods}:
We compared our anonymization method with three other anonymization approaches and compare them across the different evaluation metrics.

\textit{WordNet Model}: We implement the approach of \cite{c:63}. It assumes that since most stylometric analysis uses word frequencies to determine authorship, replacing high frequency words with lower frequency ones will help evade detection. This rule-based approach uses the Wordnet language model to replace high frequency words with their synonyms.  The method chooses the synonym based on the similarity to the original word, and the probability the new word would appear in the sentence. The word frequencies are determined using a corpus of previously written work by the author. 

\textit{Round Trip Translation}: Round trip translation takes advantage of machine translation for anonymization. We adopt the approach of \cite{c:62}, where sentences are translated from English to German to French and back to English. We use the Google Translation API to perform these translations. This method for obfuscation is not authorship-based, but rather relies on distortions and changes created by the translations to function as anonymization.

\textit{Mutant-X}: The Mutant-X method developed by \citet{c:21} alters a document by making word replacements that are close according to Word2Vec and sentiment based embedding. At each iteration of the method, copies of the document are made, each with different changes. The fitness of each copy is determined by a combination of the new document's meteor score, and its ability to fool an author classification method (for this we developed a CNN distinct from the evaluation classifiers). The best copies are then chosen for further alteration and the process continues until an iteration limit is reached, in which case the method says it cannot anonymize the sentence, or an alteration leads to successful obfuscation ( e.g., the classifier incorrectly classifies the author). By design, this approach does not alter a document if the classifier incorrectly guesses the original document, since the original document has already met the termination conditions.

\textbf{Closed-Set Results}
Results for closed-set authorship anonymization are given in Table 1. Mutant-X performs best in terms of content preservation and fluency, but along with WordNet, performs poorly when it comes to anonymizing sentences. However, our method performs comparably to the other baselines in terms of sentence perplexity and content preservation, but outperforms in terms of anonymization performance. This is because Mutant-X anonymizes based on a known adversary and iteratively makes changes to a sentence to fool that adversary. This results in minimal changes but a lack of ability to generalize when faced with previously unseen adversaries. In many cases, these methods often do not make any changes to the input sentence. This can be seen in example sentences from the test dataset:
\begin{quote}
    \textbf{Original}: all the elements that can lead to famine are nevertheless there , said [PERSON] \\
    \textbf{Wordnet}: all the components that can lead to famine are nevertheless there , said [PERSON] \\ 
    \textbf{Roundtrip}: all the elements that could lead to starvation are still there, the [PERSON] said \\
    \textbf{Mutant-X}: all the elements that can lead to famine are nevertheless there , said [PERSON] \\
    \textbf{Ours}: all the problems that can lead to famine are certainly there , said [PERSON] 
\end{quote}
Here, we can see differences between the results of our method versus the baselines. Wordnet makes some modifications to the sentence, but its modifications are generally insufficient to fool classifiers. Mutant-X is either unable to anonymize the sentence or determined that it does not need to be anonymized, which also explains why the classifier accuracy on the Mutant-X anonymized sentences is so high. While round-trip translation does change sentences, it does so simply as a result of distortions introduced in the round-trip translation rather than with any intent to anonymize the sentence. However, our method recognizes context and substitutes ``problems'' for ``elements'' and ``certainly'' for ``nevertheless''.

\textbf{Open-Set Results}
The generalizability of our method can also be seen in Table 2 with the results of open-set anonymization. Evaluated independently, our methodology performs well. On unseen authors, it drops the F1 score of classifiers trained on those authors by an average of 0.24. The model also retains fluency and content as well as its closed-set counterpart. We take this as evidence that our model, through competing with a Siamese discriminator, is better able to learn how to recognize style instead of merely removing the most recognizable features of authors that it has previously seen. To our knowledge, this is the first authorship-based anonymization approach that has been tested and proven successful in an open-set context.

However, round-trip translation performs comparably to our model in terms of F1 score, which is unexpected since round-trip translation does not intend to obfuscate authorship whereas our model does. In a closed-set context, our model has an advantage because it learns to obfuscate style through learning to anonymize sentences from authors in the test set. However, in an open-set context, despite good generalizability, this advantage is lost which means that the distortions introduced by translation are better able to compete with the anonymization from our model. For example, 
\begin{quote}
    \textbf{Original}: [ORG] , often called a vulture fund , received widespread attention last month when it made the person buy . \\
    \textbf{Round-trip}: [ORG], often referred to as the Vulture Fund, gained a lot of attention last month when he led the person to buy. \\
    \textbf{Ours}: [ORG] , often called a vulture fund , received massive attention last month when it made the person buy . 
\end{quote}
While the our model's changes are targeted towards style obfuscation, the more random changes in sentence structure and word choice created by round-trip translation occur often enough to fool the adversarial classifiers at similar rates.


\section{Conclusion}


We introduced a novel methodology for authorship obfuscation, which uses a Siamese discriminator that captures style, fluency, and content losses to provide granular feedback to the generator throughout the GAN training process. We also showed that our  model anonymizes well in an open-set setting, where it reduces the accuracy of adversarial classifiers trained on an author's sentences even when the model itself has not been trained using that author's text. However, future work remains. One issue with the anonymized output from our model is the repeated presence of certain words despite not making sense in context, which is an extension of the mode collapse problem  in GAN's. Our model learns a specific word is useful for anonymization and general enough to avoid increasing content loss. For example, we note this behavior with the word ``company'':
\begin{quote}
\textbf{Original}: [ORG] will keep 60 percent of [ORG] while the french state will have a golden share to safeguard national security interests . \\
\textbf{Ours}:  [ORG] company company 60 percent of [ORG] while the french state will have a golden taste to safeguard national security interests .
\end{quote} 
This suggests that our mechanism for content loss needs to be improved. While the FastText embeddings provide valuable feedback in terms of maintaining content similarity, they lack context. The average pooling aims to provide this context but is limited in its ability to do so. Designing a better content loss mechanism will also help lower the perplexity of sentences, since the words will be easier to predict. 

\section{Ethical Statement}
This paper provides a methodology to anonymize text with the intent of protecting free and anonymous speech, an integral part of democracy. The proposed system would enable those who are afraid to speak out against oppressive systems to do so with more confidence in their anonymity. However, there are legitimate uses of authorship attribution systems in both plagiarism detection and law enforcement. Because plagiarism often occurs within an academic context where written work is evaluated on style and expressiveness, this risk seems secondary. More importantly, law enforcement may legitimately use authorship attribution to identify criminal activity (such as identifying predators online). That informs our choice to focus on a journalistic corpus rather than one that is more representative of online text communications (or other avenues for potential criminal activity).  We also recognize that against a new, innovative method for authorship attribution, our system cannot provide any guarantees for anonymity. Despite these risks, we still view this work as progress in protecting the anonymous speech of political dissidents, journalists, and whistleblowers, which we believe outweigh the potential risks.

\bibliography{aaai22}

\begin{thebibliography}{62}
\providecommand{\natexlab}[1]{#1}

\bibitem[{Afroz, Brennan, and Greenstadt(2012)}]{c:49}
Afroz, S.; Brennan, M.; and Greenstadt, R. 2012.
\newblock Detecting Hoaxes, Frauds, and Deception in Writing Style Online.
\newblock In \emph{2012 IEEE Symposium on Security and Privacy}, 461--475.

\bibitem[{Anonymous(2018)}]{c:25}
Anonymous. 2018.
\newblock I am part of the resistance inside the trump administration.

\bibitem[{Arjovsky and Bottou(2017)}]{c:39}
Arjovsky, M.; and Bottou, L. 2017.
\newblock Towards Principled Methods for Training Generative Adversarial
  Networks.
\newblock arXiv:1701.04862.

\bibitem[{Arjovsky, Chintala, and Bottou(2017)}]{c:12}
Arjovsky, M.; Chintala, S.; and Bottou, L. 2017.
\newblock Wasserstein GAN.
\newblock arXiv:1701.07875.

\bibitem[{Bengio et~al.(2015)Bengio, Vinyals, Jaitly, and Shazeer}]{c:32}
Bengio, S.; Vinyals, O.; Jaitly, N.; and Shazeer, N. 2015.
\newblock Scheduled Sampling for Sequence Prediction with Recurrent Neural
  Networks.
\newblock \emph{CoRR}, abs/1506.03099.

\bibitem[{Bo et~al.(2019)Bo, Ding, Fung, and Iqbal}]{c:22}
Bo, H.; Ding, S. H.~H.; Fung, B. C.~M.; and Iqbal, F. 2019.
\newblock {ER-AE:} Differentially-private Text Generation for Authorship
  Anonymization.
\newblock \emph{CoRR}, abs/1907.08736.

\bibitem[{Borenstein(2018)}]{c:61}
Borenstein, S. 2018.
\newblock Close look at word choice could id anonymous nyt columnist: Word
  detectives.

\bibitem[{Bromley et~al.(1993)Bromley, Bentz, Bottou, Guyon, Lecun, Moore,
  Sackinger, and Shah}]{c:45}
Bromley, J.; Bentz, J.~W.; Bottou, L.; Guyon, I.; Lecun, Y.; Moore, C.;
  Sackinger, E.; and Shah, R. 1993.
\newblock SIGNATURE VERIFICATION USING A “SIAMESE” TIME DELAY NEURAL
  NETWORK.
\newblock \emph{International Journal of Pattern Recognition and Artificial
  Intelligence}, 07(04): 669--688.

\bibitem[{Caliskan and Greenstadt(2012)}]{c:18}
Caliskan, A.; and Greenstadt, R. 2012.
\newblock Translate Once, Translate Twice, Translate Thrice and Attribute:
  Identifying Authors and Machine Translation Tools in Translated Text.
\newblock In \emph{2012 IEEE Sixth International Conference on Semantic
  Computing}, 121--125.

\bibitem[{Daelemans(2013)}]{c:38}
Daelemans, W. 2013.
\newblock Explanation in Computational Stylometry.
\newblock In Gelbukh, A., ed., \emph{Computational Linguistics and Intelligent
  Text Processing}, 451--462. Berlin, Heidelberg: Springer Berlin Heidelberg.
\newblock ISBN 978-3-642-37256-8.

\bibitem[{Emmery, Manjavacas~Arevalo, and Chrupa{\l}a(2018)}]{c:64}
Emmery, C.; Manjavacas~Arevalo, E.; and Chrupa{\l}a, G. 2018.
\newblock Style Obfuscation by Invariance.
\newblock In \emph{Proceedings of the 27th International Conference on
  Computational Linguistics}, 984--996. Santa Fe, New Mexico, USA: Association
  for Computational Linguistics.

\bibitem[{Fedus, Goodfellow, and Dai(2018)}]{c:9}
Fedus, W.; Goodfellow, I.; and Dai, A.~M. 2018.
\newblock MaskGAN: Better Text Generation via Filling in the \_\_\_\_\_\_.
\newblock arXiv:1801.07736.

\bibitem[{Goodfellow et~al.(2014)Goodfellow, Pouget-Abadie, Mirza, Xu,
  Warde-Farley, Ozair, Courville, and Bengio}]{c:1}
Goodfellow, I.~J.; Pouget-Abadie, J.; Mirza, M.; Xu, B.; Warde-Farley, D.;
  Ozair, S.; Courville, A.; and Bengio, Y. 2014.
\newblock Generative Adversarial Networks.
\newblock arXiv:1406.2661.

\bibitem[{Gulrajani et~al.(2017)Gulrajani, Ahmed, Arjovsky, Dumoulin, and
  Courville}]{c:44}
Gulrajani, I.; Ahmed, F.; Arjovsky, M.; Dumoulin, V.; and Courville, A.~C.
  2017.
\newblock Improved Training of Wasserstein GANs.
\newblock \emph{CoRR}, abs/1704.00028.

\bibitem[{Guo et~al.(2017)Guo, Lu, Cai, Zhang, Yu, and Wang}]{c:8}
Guo, J.; Lu, S.; Cai, H.; Zhang, W.; Yu, Y.; and Wang, J. 2017.
\newblock Long Text Generation via Adversarial Training with Leaked
  Information.
\newblock \emph{CoRR}, abs/1709.08624.

\bibitem[{He et~al.(2018)He, Luo, Tian, and Zeng}]{c:47}
He, A.; Luo, C.; Tian, X.; and Zeng, W. 2018.
\newblock A Twofold Siamese Network for Real-Time Object Tracking.
\newblock \emph{CoRR}, abs/1802.08817.

\bibitem[{He et~al.(2015)He, Zhang, Ren, and Sun}]{c:43}
He, K.; Zhang, X.; Ren, S.; and Sun, J. 2015.
\newblock Deep Residual Learning for Image Recognition.
\newblock \emph{CoRR}, abs/1512.03385.

\bibitem[{Heusel et~al.(2017)Heusel, Ramsauer, Unterthiner, Nessler, Klambauer,
  and Hochreiter}]{c:50}
Heusel, M.; Ramsauer, H.; Unterthiner, T.; Nessler, B.; Klambauer, G.; and
  Hochreiter, S. 2017.
\newblock GANs Trained by a Two Time-Scale Update Rule Converge to a Nash
  Equilibrium.
\newblock \emph{CoRR}, abs/1706.08500.

\bibitem[{Honnibal and Montani(2017)}]{c:54}
Honnibal, M.; and Montani, I. 2017.
\newblock spaCy 2: Natural language understanding with Bloom embeddings,
  convolutional neural networks and incremental parsing.
\newblock \emph{To appear}.

\bibitem[{Irving(2016)}]{c:24}
Irving, M. 2016.
\newblock Computer analysis reveals Shakespeare's collaborators.

\bibitem[{Islam et~al.(2015)Islam, Yamaguchi, Dauber, Harang, Rieck,
  Greenstadt, and Narayanan}]{c:30}
Islam, A.~C.; Yamaguchi, F.; Dauber, E.; Harang, R.~E.; Rieck, K.; Greenstadt,
  R.; and Narayanan, A. 2015.
\newblock When Coding Style Survives Compilation: De-anonymizing Programmers
  from Executable Binaries.
\newblock \emph{CoRR}, abs/1512.08546.

\bibitem[{Jang, Gu, and Poole(2017)}]{c:7}
Jang, E.; Gu, S.; and Poole, B. 2017.
\newblock Categorical Reparameterization with Gumbel-Softmax.
\newblock arXiv:1611.01144.

\bibitem[{Jardine(2018)}]{c:27}
Jardine, E. 2018.
\newblock Tor, what is it good for? Political repression and the use of online
  anonymity-granting technologies.
\newblock \emph{New Media \& Society}, 20(2): 435--452.

\bibitem[{Juola(2013)}]{c:23}
Juola, P. 2013.
\newblock How a computer program helped Show J.K. Rowling write A Cuckoo's
  Calling.

\bibitem[{Keswani et~al.(2016)Keswani, Trivedi, Mehta, and Majumder}]{c:62}
Keswani, Y.; Trivedi, H.; Mehta, P.; and Majumder, P. 2016.
\newblock Author Masking through Translation.
\newblock In \emph{CLEF}.

\bibitem[{Kim(2014)}]{c:42}
Kim, Y. 2014.
\newblock Convolutional Neural Networks for Sentence Classification.
\newblock \emph{CoRR}, abs/1408.5882.

\bibitem[{Koch(2015)}]{c:48}
Koch, G.~R. 2015.
\newblock Siamese Neural Networks for One-Shot Image Recognition.
\newblock \emph{2015 International Conference on Machine Learning}.

\bibitem[{Lagutina et~al.(2019)Lagutina, Lagutina, Boychuk, Vorontsova,
  Shliakhtina, Belyaeva, Paramonov, and Demidov}]{c:35}
Lagutina, K.; Lagutina, N.; Boychuk, E.; Vorontsova, I.; Shliakhtina, E.;
  Belyaeva, O.; Paramonov, I.; and Demidov, P. 2019.
\newblock A Survey on Stylometric Text Features.
\newblock In \emph{2019 25th Conference of Open Innovations Association
  (FRUCT)}, 184--195.

\bibitem[{Lewis et~al.(2004)Lewis, Yang, Rose, and Li}]{c:52}
Lewis, D.~D.; Yang, Y.; Rose, T.~G.; and Li, F. 2004.
\newblock RCV1: A New Benchmark Collection for Text Categorization Research.
\newblock \emph{J. Mach. Learn. Res.}, 5: 361–397.

\bibitem[{Liu(2011)}]{c:53}
Liu, Z. 2011.
\newblock Reuter\_50\_50 Data Set.

\bibitem[{Maddison, Mnih, and Teh(2016)}]{c:6}
Maddison, C.~J.; Mnih, A.; and Teh, Y.~W. 2016.
\newblock The Concrete Distribution: {A} Continuous Relaxation of Discrete
  Random Variables.
\newblock \emph{CoRR}, abs/1611.00712.

\bibitem[{Mahlberg(2013)}]{c:34}
Mahlberg, M. 2013.
\newblock Corpus Stylistics and Dickens's Fiction.
\newblock \emph{Corpus Stylistics and Dickens's Fiction}, 1--222.

\bibitem[{Mahmood et~al.(2019)Mahmood, Ahmad, Shafiq, Srinivasan, and
  Zaffar}]{c:21}
Mahmood, A.; Ahmad, F.; Shafiq, Z.; Srinivasan, P.; and Zaffar, F. 2019.
\newblock A Girl Has No Name: Automated Authorship Obfuscation using Mutant-X.
\newblock \emph{Proceedings on Privacy Enhancing Technologies}, 2019: 54 -- 71.

\bibitem[{Mansoorizadeh et~al.(2016)Mansoorizadeh, Rahgooy, Aminian, and
  Eskandari}]{c:63}
Mansoorizadeh, M.; Rahgooy, T.; Aminian, M.; and Eskandari, M. 2016.
\newblock Author Obfuscation using WordNet and Language Models.
\newblock In \emph{CLEF}.

\bibitem[{McDonald et~al.(2012)McDonald, Afroz, Caliskan, Stolerman, and
  Greenstadt}]{c:17}
McDonald, A. W.~E.; Afroz, S.; Caliskan, A.; Stolerman, A.; and Greenstadt, R.
  2012.
\newblock Use Fewer Instances of the Letter ``i'': Toward Writing Style
  Anonymization.
\newblock In Fischer-H{\"u}bner, S.; and Wright, M., eds., \emph{Privacy
  Enhancing Technologies}, 299--318. Berlin, Heidelberg: Springer Berlin
  Heidelberg.
\newblock ISBN 978-3-642-31680-7.

\bibitem[{Mikolov et~al.(2017)Mikolov, Grave, Bojanowski, Puhrsch, and
  Joulin}]{c:41}
Mikolov, T.; Grave, E.; Bojanowski, P.; Puhrsch, C.; and Joulin, A. 2017.
\newblock Advances in Pre-Training Distributed Word Representations.
\newblock \emph{CoRR}, abs/1712.09405.

\bibitem[{Mosteller and Wallace(1963)}]{c:28}
Mosteller, F.; and Wallace, D.~L. 1963.
\newblock Inference in an Authorship Problem.
\newblock \emph{Journal of the American Statistical Association}, 58(302):
  275--309.

\bibitem[{Narayanan et~al.(2012)Narayanan, Paskov, Gong, Bethencourt, Stefanov,
  Shin, and Song}]{c:29}
Narayanan, A.; Paskov, H.; Gong, N.~Z.; Bethencourt, J.; Stefanov, E.; Shin, E.
  C.~R.; and Song, D. 2012.
\newblock On the Feasibility of Internet-Scale Author Identification.
\newblock In \emph{2012 IEEE Symposium on Security and Privacy}, 300--314.

\bibitem[{Odena, Olah, and Shlens(2017)}]{c:2}
Odena, A.; Olah, C.; and Shlens, J. 2017.
\newblock Conditional Image Synthesis With Auxiliary Classifier GANs.
\newblock arXiv:1610.09585.

\bibitem[{Papineni et~al.(2002)Papineni, Roukos, Ward, and Zhu}]{c:56}
Papineni, K.; Roukos, S.; Ward, T.; and Zhu, W.-J. 2002.
\newblock BLEU: A Method for Automatic Evaluation of Machine Translation.
\newblock In \emph{Proceedings of the 40th Annual Meeting on Association for
  Computational Linguistics}, ACL '02, 311–318. USA: Association for
  Computational Linguistics.

\bibitem[{Perlroth(2012)}]{c:36}
Perlroth, N. 2012.
\newblock Software helps identify anonymous writers or helps them stay that
  way.

\bibitem[{Radford et~al.(2019)Radford, Wu, Child, Luan, Amodei, and
  Sutskever}]{c:55}
Radford, A.; Wu, J.; Child, R.; Luan, D.; Amodei, D.; and Sutskever, I. 2019.
\newblock Language Models are Unsupervised Multitask Learners.
\newblock \emph{Preprint}.

\bibitem[{Rao and Rohatgi(2000)}]{c:19}
Rao, J.~R.; and Rohatgi, P. 2000.
\newblock Can Pseudonymity Really Guarantee Privacy?
\newblock In \emph{9th {USENIX} Security Symposium ({USENIX} Security 00)}.
  Denver, CO: {USENIX} Association.

\bibitem[{Reiter(2018)}]{c:58}
Reiter, E. 2018.
\newblock A Structured Review of the Validity of {BLEU}.
\newblock \emph{Computational Linguistics}, 44(3): 393--401.

\bibitem[{Ruder, Ghaffari, and Breslin(2016)}]{c:14}
Ruder, S.; Ghaffari, P.; and Breslin, J.~G. 2016.
\newblock Character-level and Multi-channel Convolutional Neural Networks for
  Large-scale Authorship Attribution.
\newblock \emph{CoRR}, abs/1609.06686.

\bibitem[{Saedi and Dras(2019)}]{c:15}
Saedi, C.; and Dras, M. 2019.
\newblock Siamese Networks for Large-Scale Author Identification.
\newblock \emph{CoRR}, abs/1912.10616.

\bibitem[{Salimans et~al.(2016)Salimans, Goodfellow, Zaremba, Cheung, Radford,
  and Chen}]{c:11}
Salimans, T.; Goodfellow, I.~J.; Zaremba, W.; Cheung, V.; Radford, A.; and
  Chen, X. 2016.
\newblock Improved Techniques for Training GANs.
\newblock \emph{CoRR}, abs/1606.03498.

\bibitem[{Schuster and Nakajima(2012)}]{c:51}
Schuster, M.; and Nakajima, K. 2012.
\newblock Japanese and Korean voice search.
\newblock In \emph{2012 IEEE International Conference on Acoustics, Speech and
  Signal Processing (ICASSP)}, 5149--5152.

\bibitem[{Shetty, Schiele, and Fritz(2017)}]{c:20}
Shetty, R.; Schiele, B.; and Fritz, M. 2017.
\newblock A\({}^{\mbox{4}}\)NT: Author Attribute Anonymity by Adversarial
  Training of Neural Machine Translation.
\newblock \emph{CoRR}, abs/1711.01921.

\bibitem[{Sulem, Abend, and Rappoport(2018)}]{c:59}
Sulem, E.; Abend, O.; and Rappoport, A. 2018.
\newblock {BLEU} is Not Suitable for the Evaluation of Text Simplification.
\newblock In \emph{Proceedings of the 2018 Conference on Empirical Methods in
  Natural Language Processing}, 738--744. Brussels, Belgium: Association for
  Computational Linguistics.

\bibitem[{Sundararajan(2018)}]{c:37}
Sundararajan, K. 2018.
\newblock Analysis of Stylometry as a Cognitive Biometric Trait.

\bibitem[{Taigman et~al.(2014)Taigman, Yang, Ranzato, and Wolf}]{c:46}
Taigman, Y.; Yang, M.; Ranzato, M.; and Wolf, L. 2014.
\newblock DeepFace: Closing the Gap to Human-Level Performance in Face
  Verification.
\newblock In \emph{2014 IEEE Conference on Computer Vision and Pattern
  Recognition}, 1701--1708.

\bibitem[{Turian, Shen, and Melamed(2003)}]{c:57}
Turian, J.; Shen, L.; and Melamed, I. 2003.
\newblock Evaluation of Machine Translation and its Evaluation.
\newblock \emph{2003 Machine Translation Summit}.

\bibitem[{Vaswani et~al.(2017)Vaswani, Shazeer, Parmar, Uszkoreit, Jones,
  Gomez, Kaiser, and Polosukhin}]{c:31}
Vaswani, A.; Shazeer, N.; Parmar, N.; Uszkoreit, J.; Jones, L.; Gomez, A.~N.;
  Kaiser, L.; and Polosukhin, I. 2017.
\newblock Attention Is All You Need.
\newblock \emph{CoRR}, abs/1706.03762.

\bibitem[{Williams(2017)}]{c:26}
Williams, J. 2017.
\newblock To protect our democracy, we need to protect anonymous low-cost
  online political speech.

\bibitem[{Williams(2004)}]{c:5}
Williams, R.~J. 2004.
\newblock Simple statistical gradient-following algorithms for connectionist
  reinforcement learning.
\newblock \emph{Machine Learning}, 8: 229--256.

\bibitem[{Wu et~al.(2016)Wu, Zhang, Xue, Freeman, and Tenenbaum}]{c:3}
Wu, J.; Zhang, C.; Xue, T.; Freeman, W.~T.; and Tenenbaum, J.~B. 2016.
\newblock Learning a Probabilistic Latent Space of Object Shapes via 3D
  Generative-Adversarial Modeling.
\newblock \emph{CoRR}, abs/1610.07584.

\bibitem[{Yang et~al.(2018)Yang, Hu, Dyer, Xing, and Berg{-}Kirkpatrick}]{c:40}
Yang, Z.; Hu, Z.; Dyer, C.; Xing, E.~P.; and Berg{-}Kirkpatrick, T. 2018.
\newblock Unsupervised Text Style Transfer using Language Models as
  Discriminators.
\newblock \emph{CoRR}, abs/1805.11749.

\bibitem[{Yu et~al.(2016)Yu, Zhang, Wang, and Yu}]{c:10}
Yu, L.; Zhang, W.; Wang, J.; and Yu, Y. 2016.
\newblock SeqGAN: Sequence Generative Adversarial Nets with Policy Gradient.
\newblock \emph{CoRR}, abs/1609.05473.

\bibitem[{Zhang et~al.(2019{\natexlab{a}})Zhang, Kishore, Wu, Weinberger, and
  Artzi}]{c:60}
Zhang, T.; Kishore, V.; Wu, F.; Weinberger, K.~Q.; and Artzi, Y.
  2019{\natexlab{a}}.
\newblock BERTScore: Evaluating Text Generation with {BERT}.
\newblock \emph{CoRR}, abs/1904.09675.

\bibitem[{Zhang et~al.(2019{\natexlab{b}})Zhang, Feng, Meng, You, and
  Liu}]{c:33}
Zhang, W.; Feng, Y.; Meng, F.; You, D.; and Liu, Q. 2019{\natexlab{b}}.
\newblock Bridging the Gap between Training and Inference for Neural Machine
  Translation.
\newblock \emph{CoRR}, abs/1906.02448.

\bibitem[{Zhang, Cai, and Zhang(2017)}]{c:4}
Zhang, Y.; Cai, W.; and Zhang, Y. 2017.
\newblock Separating Style and Content for Generalized Style Transfer.
\newblock \emph{CoRR}, abs/1711.06454.

\end{thebibliography}


\end{document}